\documentclass[conference]{IEEEtran}
\IEEEoverridecommandlockouts
\usepackage{cite}
\usepackage{amsmath,amssymb,amsfonts}
\usepackage{algorithmic}
\usepackage{graphicx}
\usepackage{textcomp}
\usepackage{xcolor}
\usepackage{subcaption}
\def\BibTeX{{\rm B\kern-.05em{\sc i\kern-.025em b}\kern-.08em
    T\kern-.1667em\lower.7ex\hbox{E}\kern-.125emX}}
\begin{document}

\title{Synthetic Feature Augmentation Improves Generalization Performance of Language Models
}

\author{\parbox{16cm}{\centering
    {\large Ashok Choudhary$^{1}$, Cornelius Thiels$^{1}$, Hojjat Salehinejad$^{2,3}$,~\IEEEmembership{Senior Member,~IEEE}\\
    {\normalsize
    $^1$Department of Surgery, Mayo Clinic, Rochester, MN, USA \\    
    $^2$Kern Center for the Science of Health Care Delivery, Mayo Clinic, Rochester, MN, USA \\
    $^3$Department of Artificial Intelligence and Informatics, Mayo Clinic, Rochester, MN, USA \\    
    \textit{\{choudhary.ashok, thiels.cornelius, salehinejad.hojjat\}@mayo.edu}
    }}
}}

\maketitle

\begin{abstract}
Training and fine-tuning deep learning models, especially large language models (LLMs), on limited and imbalanced datasets poses substantial challenges. These issues often result in poor generalization, where models overfit to dominant classes and underperform on minority classes, leading to biased predictions and reduced robustness in real-world applications. To overcome these challenges, we propose augmenting features in the embedding space by generating synthetic samples using a range of techniques. By upsampling underrepresented classes, this method improves model performance and alleviates data imbalance. We validate the effectiveness of this approach across multiple open-source text classification benchmarks, demonstrating its potential to enhance model robustness and generalization in imbalanced data scenarios.
\end{abstract}


\section{Introduction}
Language models are computational frameworks designed to understand and generate human language by analyzing textual data to capture patterns and structures. Advanced models, such as bidirectional encoder representations from transformers (BERT)~\cite{devlin2019bert} and generative pre-trained transformer 4 (GPT-4), leverage deep learning and large datasets to excel at natural language processing tasks like text generation, translation, and sentiment analysis.

Fine-tuning large language models (LLMs) on limited-imbalanced datasets~\cite{rengers2024academic} is challenging, often leading to overfitting and biased predictions~\cite{choi2024dataaug}. For instance, in healthcare, models trained on datasets with underrepresented rare diseases may overlook these cases, reducing diagnostic accuracy. This issue is prevalent across domains where class imbalance skews model performance toward majority classes.

To overcome above challenges, methods like data augmentation~\cite{mialon2023augmentedlanguagemodelssurvey}, transfer learning~\cite{10.5555/3455716.3455856}, synthetic minority over-sampling technique (SMOTE)~\cite{Chawla_2002}, variational autoencoders (VAEs)~\cite{Chawla2002}, and data synthesizing~\cite{salehinejad2018generalization} can be used. Transfer learning allows pre-trained models to adapt to specific domains, while ensemble approaches enhance prediction stability. These techniques enable more robust, equitable models capable of handling real-world data complexities.

In this study, we explore the use of synthetic data generation within the embedding spaces of pre-trained models, such as BERT, to address model bias caused by class imbalance. By leveraging these embedding spaces, we employ advanced data augmentation techniques, including SMOTE and VAEs, to generate synthetic samples that closely represent the original data distributions. Our results show that incorporating these synthetic samples into imbalanced training datasets significantly enhances classification performance compared to training without synthetic augmentation. We assess the effectiveness of these methods on various benchmark datasets, demonstrating their potential to improve model robustness and fairness in real-world applications.

\begin{figure*}[!t]
\captionsetup{font=footnotesize}
\centering
\includegraphics[width=\textwidth, trim=50 1250 10 20, clip]{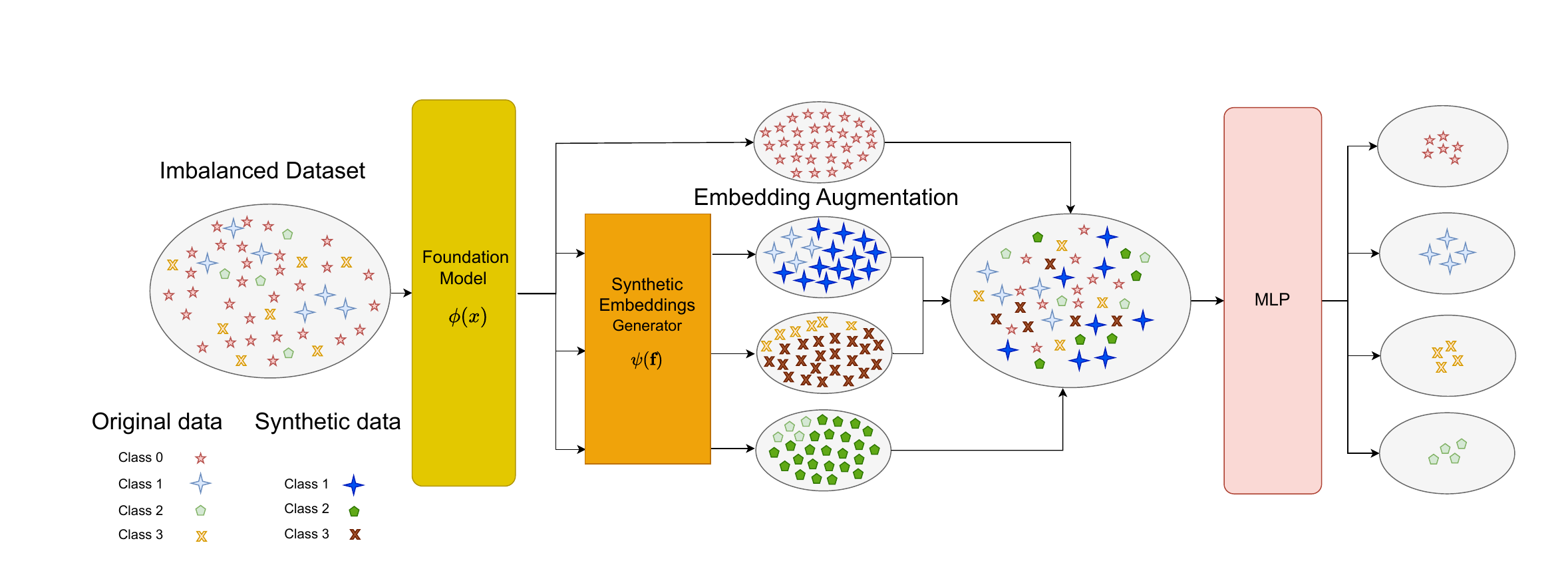}
\caption{Multiclass classification model addressing imbalance by generating synthetic samples to mitigate bias from the limited, imbalanced dataset.}
\label{fig:dataaugmentation}
\vspace{-6mm}
\end{figure*}

\section{Embedding Space Augmentation Models}

Figure~\ref{fig:dataaugmentation} shows different steps of the proposed method. Let $\mathcal{D} = \{ (x_1, y_1), \dots, (x_n, y_n) \}$ be a dataset of  text samples $x_i$ and corresponding labels $y_i$. A nonlinear embedding function maps $x_i$ to a $d$-dimensional vector $\mathbf{f}_i$ as  
\begin{equation}
 \mathbf{f}_i = \mathbf{\phi}(x_i),   
 \label{eq:embedding}
\end{equation}
where it captures intricate semantic and syntactic relationships within the text by mapping words or tokens into continuous numerical spaces. The generated embeddings are then passed to a synthetic embedding generator $\psi(\mathbf{f_i})$, which synthesizes embedding vectors for the minority data class. The synthesized embedding vectors are then combined with the real embedding vectors to form a balanced training dataset. This balanced dataset is then used to train a classifier~\cite{salehinejad2018generalization,salehinejad2018synthesizing}.


 \begin{figure}[!t]
\captionsetup{font=footnotesize}
\centering
\includegraphics[width=\textwidth, trim=50 1450 100 80, clip]{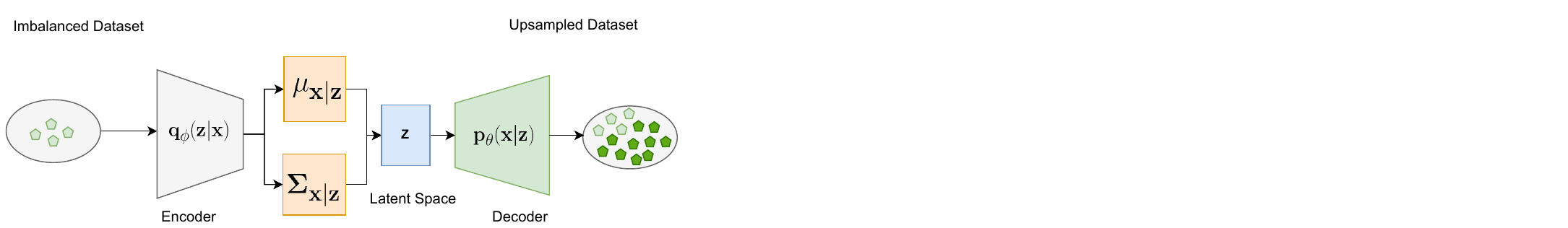}
\caption{Variational autoencoder model applied to generate synthetic features.}
\label{fig:VAE}
\vspace{-6mm}
\end{figure}

\begin{figure*}[!t]
\captionsetup{font=footnotesize}
    \centering
        \centering
        \begin{subfigure}[b]{0.32\textwidth}
    \includegraphics[width=\textwidth,trim=55 75 55 50,clip]{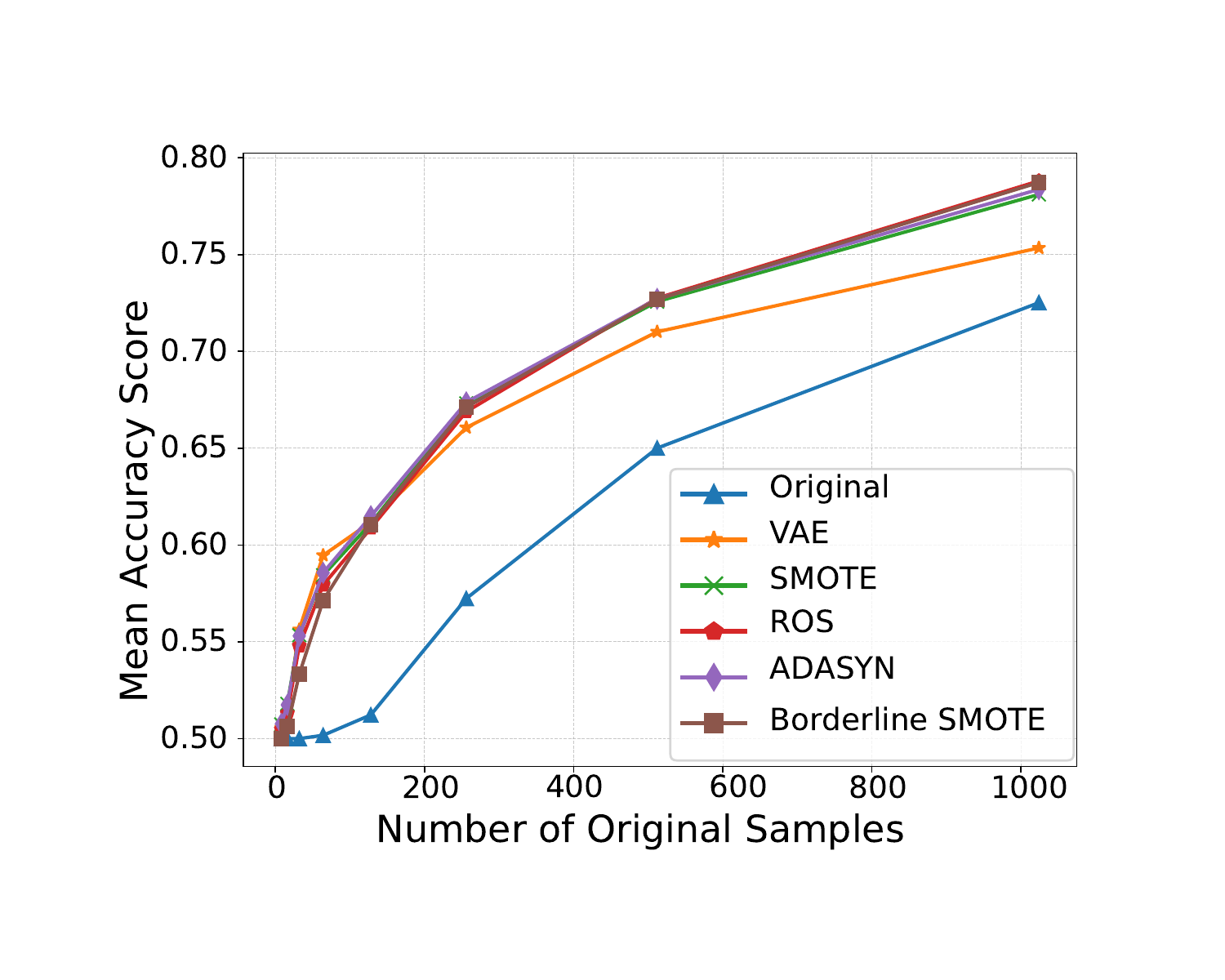}        
        \vspace{-4mm}
        \caption{SST-2 dataset.}
        \label{fig:sst2_datasets_accuracy_plots}
        \end{subfigure}                
    \begin{subfigure}[b]{0.32\textwidth}
        \centering            \includegraphics[width=\textwidth,trim=55 75 55 50,clip]{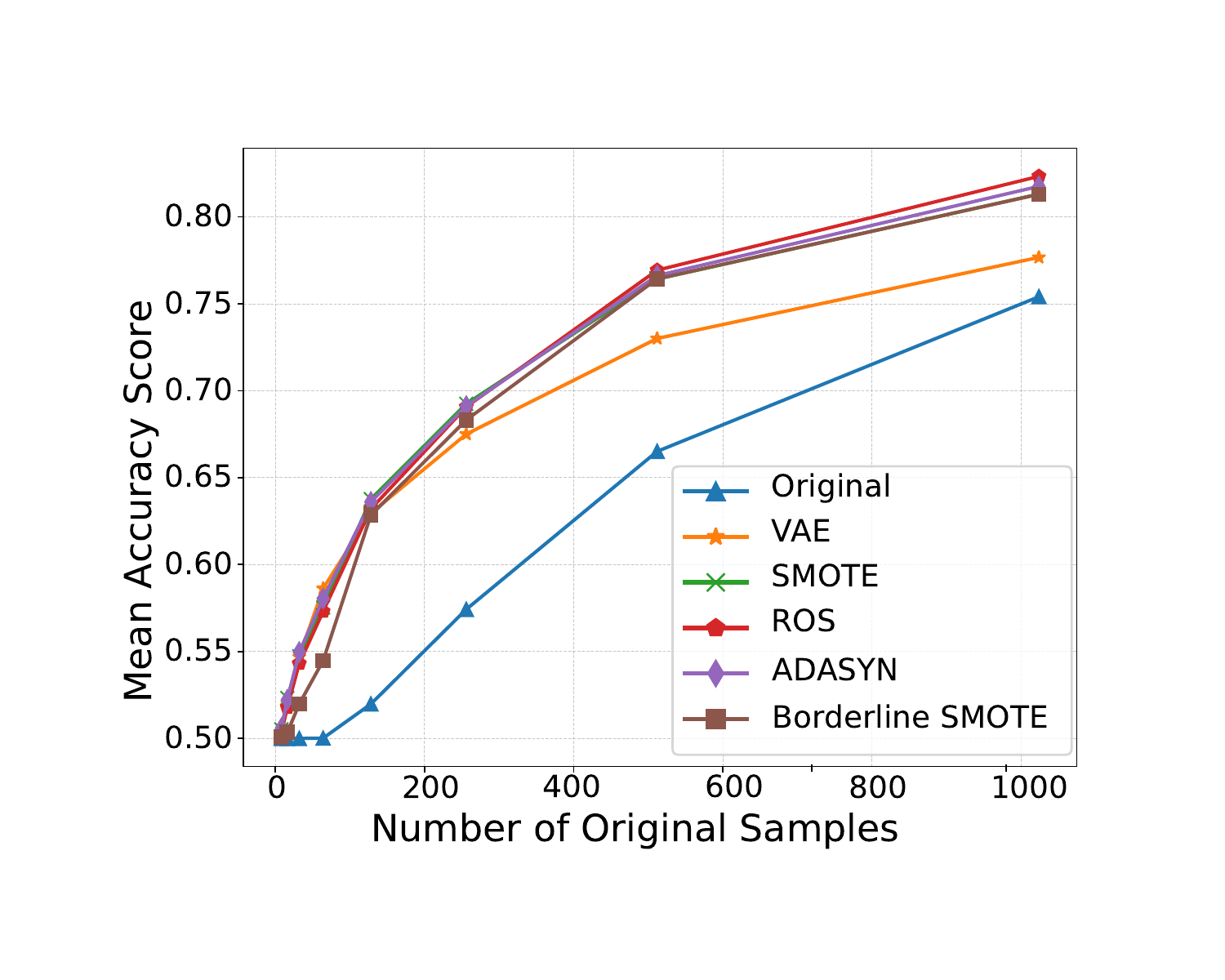}
        \vspace{-4mm}
        \caption{IMDB dataset.}
        \label{fig:imdb_datasets_accuracy_plots}
    \end{subfigure}
    \caption{Performance results on the SST-2 and IMDB datasets.}
    \label{fig:sst_imdb_results}
    \vspace{-4mm}
\end{figure*}

\begin{figure*}[t]
\captionsetup{font=footnotesize}
\centering
    \begin{subfigure}[t]{0.34\textwidth}
        \includegraphics[width=\textwidth, trim=50 70 100 80, clip]{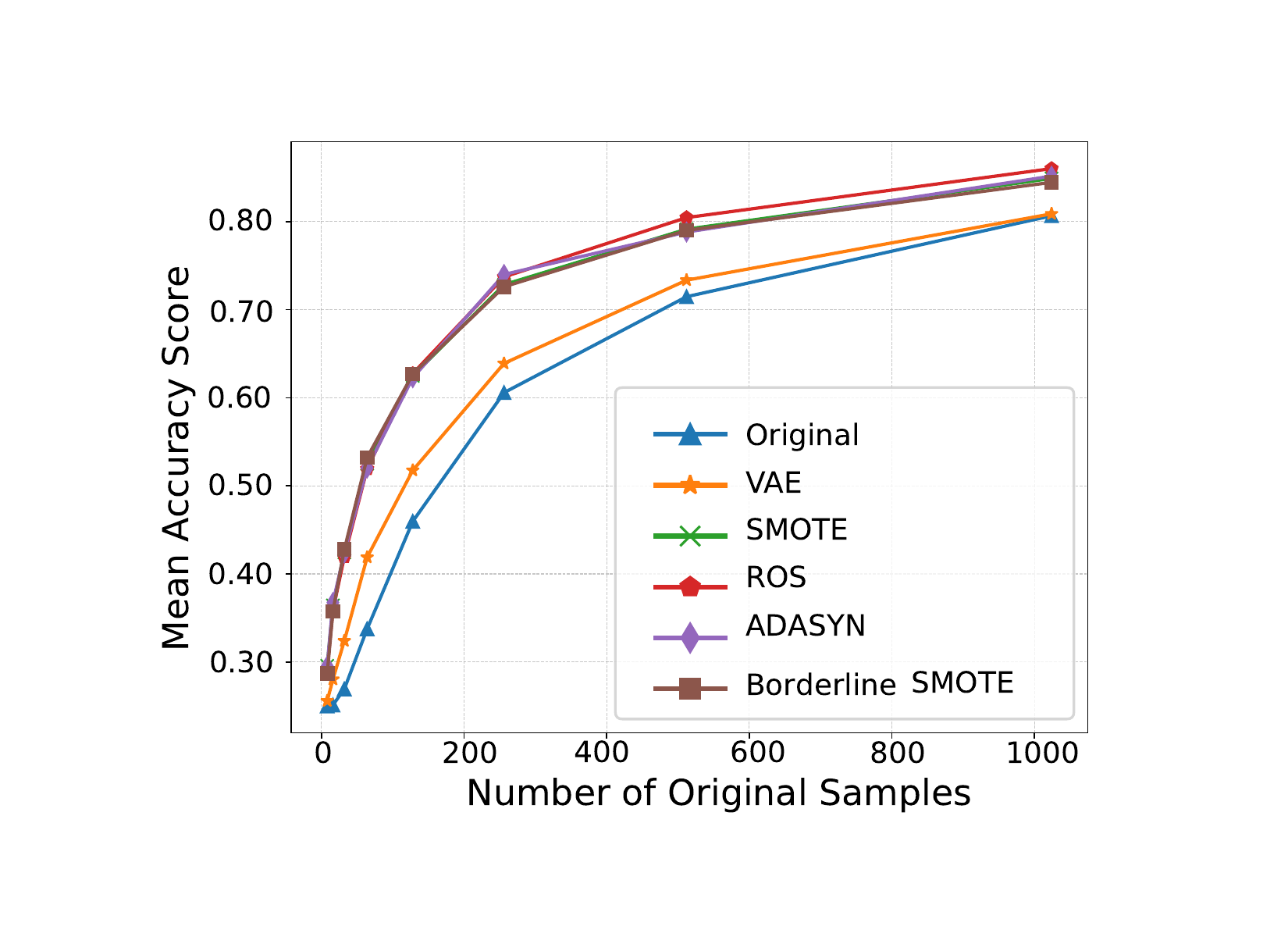}
           \vspace{-4mm}
        \caption{Downsample labels 1, 2, and 3.}
        \label{fig:sub1}
    \end{subfigure}
    \begin{subfigure}[t]{0.32\textwidth}
        \includegraphics[width=\textwidth, trim=50 70 60 50, clip]{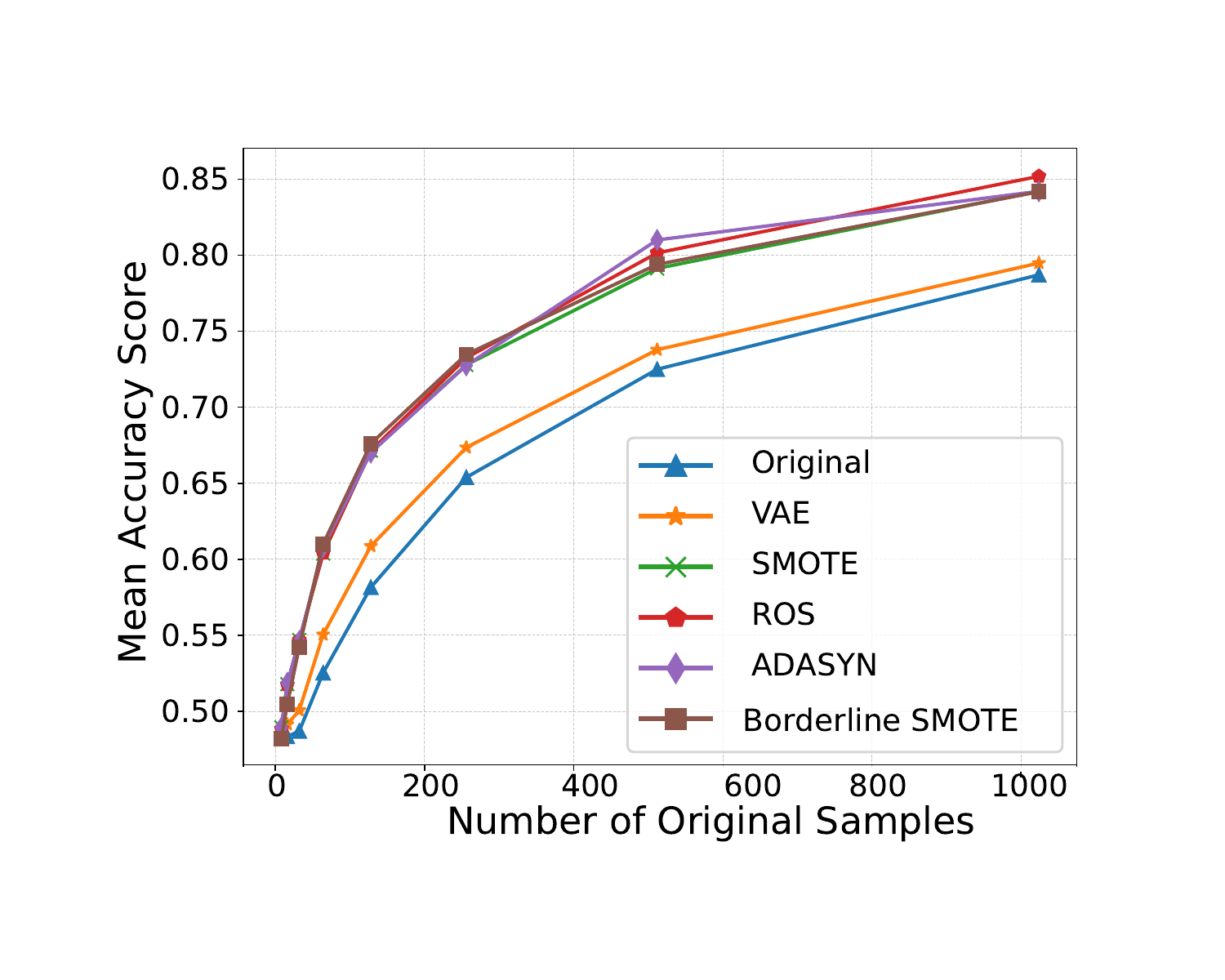}
           \vspace{-4mm}
        \caption{Downsample labels 1 and 3.}
        \label{fig:sub2}
    \end{subfigure}
    \begin{subfigure}[t]{0.32\textwidth}
        \includegraphics[width=\textwidth, trim=50 70 60 50, clip]{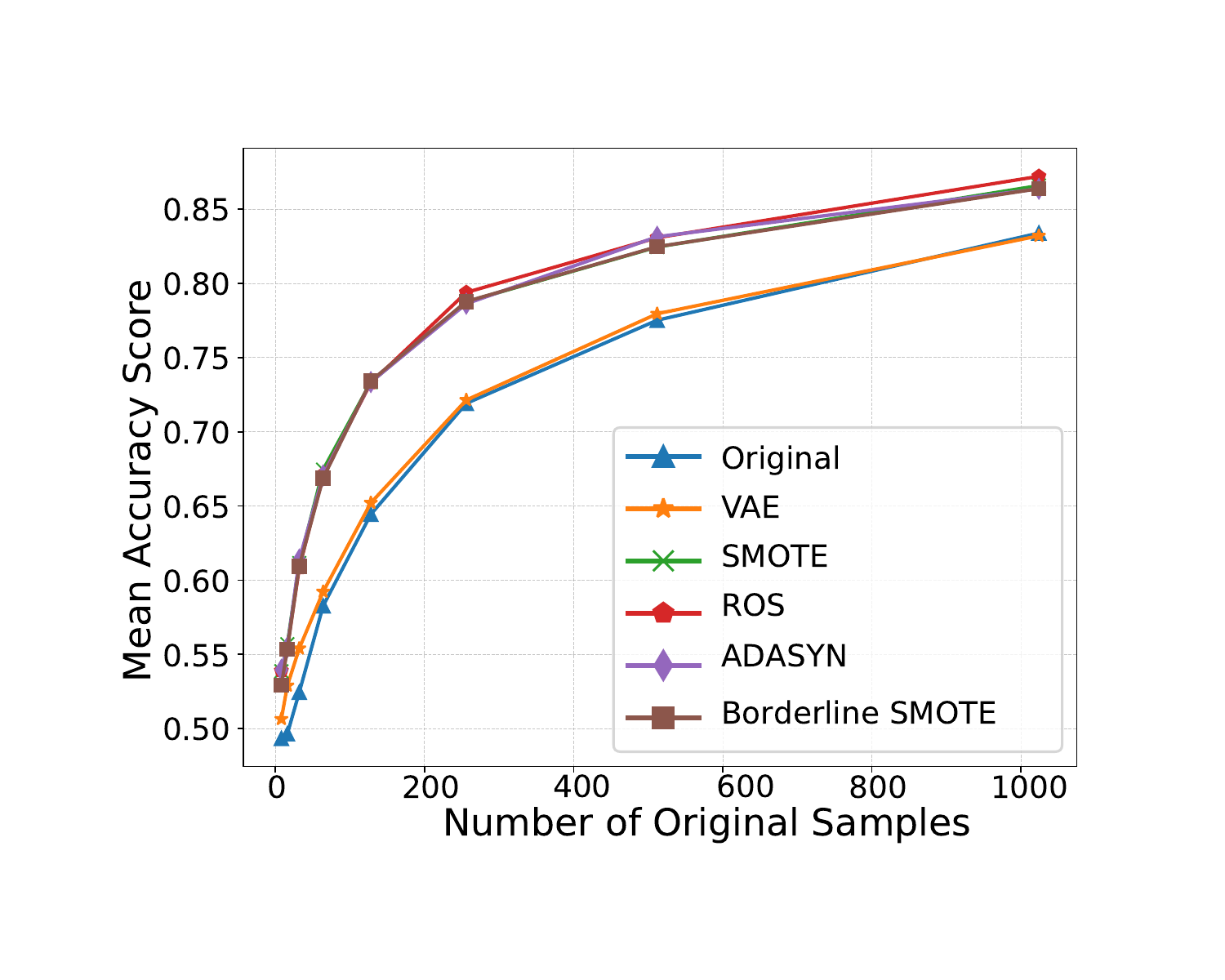}
           \vspace{-4mm}
        \caption{Downsample labels 2 and 3.}
        \label{fig:sub3}
    \end{subfigure}
        \caption{Accuracy (out of 1) vs. number of samples for the AG News dataset with different downsampling strategies.}
        \vspace{-4mm}
        \label{fig:agnews_results}
\end{figure*}

\subsection{Synthetic Minority Over-sampling Technique}
SMOTE is a popular method for addressing class imbalance by generating synthetic minority class samples. In embedding space, it captures semantic and syntactic relationships encoded in the embeddings. The technique identifies the $k$-nearest neighbors of each minority class sample in the embedding space. For a query vector $\mathbf{f}_j \in \mathcal{F}$, the goal is to find the nearest neighbor vector $\mathbf{f}_{nn} \in \mathcal{F}$ s.t. $\mathbf{f}_{nn}\neq \mathbf{f}_j$ as
\begin{equation}
\mathbf{f}_{nn} = \underset{\mathbf{f}_i \in \mathcal{F}, \mathbf{f}_i\neq \mathbf{f}_i}{\arg\min}\delta(\mathbf{f}_i, \mathbf{f}_j),
\end{equation}
where $\delta(\mathbf{f}_i, \mathbf{f}_j)$ is the distance between $\mathbf{f}_j$ and $\mathbf{f}_i$. 
For a given minority class embedding $\textbf{f}_i \in \mathbb{R}^d$, a synthetic sample $\textbf{f}_{\text{new}}$ is generated along the line segment connecting \(\textbf{f}_i\) and one of its randomly selected $k$-nearest neighbors $\textbf{f}_{i,\text{nn}} \in \mathbb{R}^d$ as
\begin{equation}
\textbf{f}_{\text{new}} = \textbf{f}_i + \lambda \left( \textbf{f}_{i,\text{nn}} - \textbf{f}_i \right),
\label{smoteeqn}
\end{equation}
where $\lambda \sim \mathcal{U}(0, 1)$ is a random scalar drawn from a uniform distribution. The synthetic sample $\textbf{f}_{\text{new}}$ lies on the straight line between the original minority class sample $\textbf{f}_i$ and its selected nearest neighbor $\textbf{f}_{i,\text{nn}}$, effectively interpolating between the two points. 

\subsection{Borderline Synthetic Minority Over-sampling Technique}
This variation of SMOTE generates synthetic samples near the decision boundary between the majority and minority classes~\cite{Han_Hui}. 
For each minority class embedding $\mathbf{f}_i \in \mathcal{F}_{\text{min}}$, where $\mathcal{F}_{\text{min}}$ is the set of minority embeddings, its $k$-nearest neighbors from both the minority and majority classes are identified in the embedding space. The sample $\mathbf{f}_i$ is considered a borderline example if most of its neighbors belong to the majority class. If $\mathbf{f}_i$ is a borderline sample, new synthetic samples are generated between $\mathbf{f}_i$ and one of its minority class neighbors $\mathbf{f}_{i,\text{nn}}$ using Eq. (\ref{smoteeqn}). This ensures that the synthetic samples are generated close to the borderline minority embeddings, improving the classifier's ability to correctly identify the decision boundary. By focusing on borderline examples, it aims to increase the classification accuracy for the minority class while reducing the risk of generating noisy or redundant samples from the minority class core.

\subsection{Adaptive Synthetic Sampling}
The adaptive synthetic sampling (ADASYN)~\cite{Haibo_He} generates synthetic samples for the minority class based on the difficulty of learning those samples. Given a dataset $\mathcal{F}$ with majority class $\mathcal{F}_{\text{maj}}$ and minority class $\mathcal{F}_{\text{min}}$, where $|\mathcal{F}_{\text{maj}}| = N_{\text{maj}}$ and $|\mathcal{F}_{\text{min}}| = N_{\text{min}}$ such that $N_{\text{maj}} > N_{\text{min}}$, the goal is to generate synthetic samples to balance the class distribution.

First, for each minority class embedding $\mathbf{f}_i \in \mathcal{F}_{\text{min}}$, the number of nearest neighbors $k_i$ from the majority class in the embedding space is computed. A difficulty score $r_i$ is then calculated for each minority class embedding as
\begin{equation}
r_i = \frac{k_i}{k_{\text{total}}},    
\end{equation}
where $k_{\text{total}}$ is the total number of nearest neighbors considered. The number of synthetic samples to generate for each minority class embedding is proportional to its difficulty score, defined as,
\begin{equation}
G_i = r_i \times G_{\text{total}},
\end{equation}
where $G_{\text{total}}$ is the total number of synthetic samples required, given by $G_{\text{total}} = N_{\text{maj}} - N_{\text{min}}$. New synthetic samples are generated using Eq. (\ref{smoteeqn}). This process results in an adaptive number of synthetic samples for harder-to-learn minority class embeddings, helping to balance the class distribution while focusing on challenging regions of the embedding space.

\subsection{Random Oversampling}
Random Oversampling (ROS)~\cite{ROS} is a technique used to address class imbalance by replicating samples from the minority class. The goal is to increase the size of the minority class by sampling with replacement to ${N_{\text{oversample}} = N_{\text{maj}} - N_{\text{min}}}$.
Assuming $\mathbf{f}_i$ represent a randomly chosen embedding from $\mathcal{F}_{\text{min}}$, where \( i = 1, 2, \dots, N_{\text{oversample}} \), the new dataset after oversampling becomes
\begin{equation}
\mathcal{F}' = \mathcal{F}_{\text{maj}} \cup \left( \mathcal{F}_{\text{min}} \cup \{ \mathbf{f}_1, \mathbf{f}_2, \dots, \mathbf{f}_{N_{\text{oversample}}} \} \right),
\end{equation}
resulting in a balanced dataset where $|\mathcal{F}'_{\text{maj}}| = |\mathcal{F}'_{\text{min}}| = N_{\text{maj}}$.

\subsection{Variational Autoencoders}
The VAEs are generative models that can be utilized to generate synthetic data by learning a probabilistic latent space representation of the input data. In this work, VAEs can be used as the embeddings generator $\phi(\cdot)$, Figure~\ref{fig:VAE}.
Given the set of embedding vectors $\mathbf{f}_i \in \mathcal{F}$, VAEs aim to learn a low-dimensional latent representation $z$ by modeling the joint distribution $p_\alpha(\mathbf{f}, z)$, where $\alpha$ represents the parameters of the model. The joint distribution is defined through the likelihood of the observed data given the latent variables, $p_\alpha(\mathbf{f}|z)$, and a prior distribution over the latent variables, $p(z)$. Training VAEs involves maximizing the Variational Evidence Lower Bound (ELBO) on the marginal likelihood \(p_\alpha(\mathbf{f})\), which serves as a computationally feasible surrogate for the true marginal likelihood. The ELBO is given by
\begin{equation}
\text{ELBO} = \mathbb{E}_{z \sim q_\beta(z|\mathbf{f})} \left[ \log p_\alpha(\mathbf{f}|z) \right] - \text{KL}\left( q_\beta(z|\mathbf{f}) \| p(z) \right),    
\end{equation}
where $q_\beta(z|\mathbf{f})$ is the variational posterior (encoder) parameterized by $\beta$, approximating the true posterior distribution of the latent variables given the data. The term \(\text{KL}\left( q_\beta(z|\mathbf{f}) \| p(z) \right)\) represents the Kullback-Leibler divergence between the variational posterior and the prior distribution over the latent variables.

By maximizing the ELBO, the VAE effectively balances the reconstruction accuracy of the input embeddings \(\mathbf{f}_i\) and the regularization imposed by the latent space. This results in a compact and meaningful representation of the embeddings. Once the VAE is trained, we can generate new synthetic embeddings by sampling from the latent space and passing these samples through the decoder network \(p_\alpha(\mathbf{f}|z)\). These synthetic embeddings are then used to augment the minority class in the embedding space, addressing class imbalance in the dataset.

\section{Experiments}
 We use publicly available datasets with binary and multi-class labels, deliberately down-sampled to create imbalanced versions. Synthetic data is generated using the discussed techniques to balance the training set. Performance is evaluated on a balanced test set using 10-fold cross-validation, with datasets split into $80\%$ training, $10\%$ validation, and $10\%$ test. We compare model performance on the original imbalanced datasets and the balanced datasets with synthetic data. For the classification task, we employed an MLP with a single hidden layer comprising $128$ hidden units. This choice of model architecture was intended to balance complexity and performance, providing a robust framework for our comparisons.

\subsection{Datasets}
\textbf{IMDB Dataset}:
The IMDB dataset~\cite{maas-etal-2011-learning} contains 50,000 movie reviews evenly split into 25,000 positive and 25,000 negative reviews, making it ideal for binary sentiment classification. The reviews are preprocessed and substantial in length, supporting detailed sentiment analysis and modeling.

\textbf{SST-2 Dataset}:
The SST-2 dataset~\cite{socher-etal-2013-recursive}, a subset of the Stanford Sentiment Treebank, is designed for binary sentiment classification at the sentence level using movie reviews from Rotten Tomatoes. 

\textbf{AG News Dataset}
The AG News dataset~\cite{zhang2016character} includes over 1 million news articles categorized into four classes: World, Sports, Business, and Science/Technology, with 120,000 training and 7,600 test samples. 

\subsection{Results}
Performance of the models were compared on imbalanced datasets versus those augmented with synthetic data. For minority classes, we systematically evaluated sample sizes in powers of two ($2^m$, where $2 < m \leq 10$) to assess their impact on performance.
Using techniques like SMOTE and VAE, models trained with SMOTE-augmented data consistently achieved higher accuracy on balanced test datasets. Figures~\ref{fig:sst_imdb_results} and~\ref{fig:agnews_results}  illustrate the performance of different embedding augmentation methods in classification of the text samples in the IMDB, SST-2, and AG News datasets.

\begin{figure}[t]
\captionsetup{font=footnotesize}
\centering
    \begin{subfigure}[t]{0.48\columnwidth} 
        \includegraphics[width=\textwidth, trim=80 60 115 50, clip]{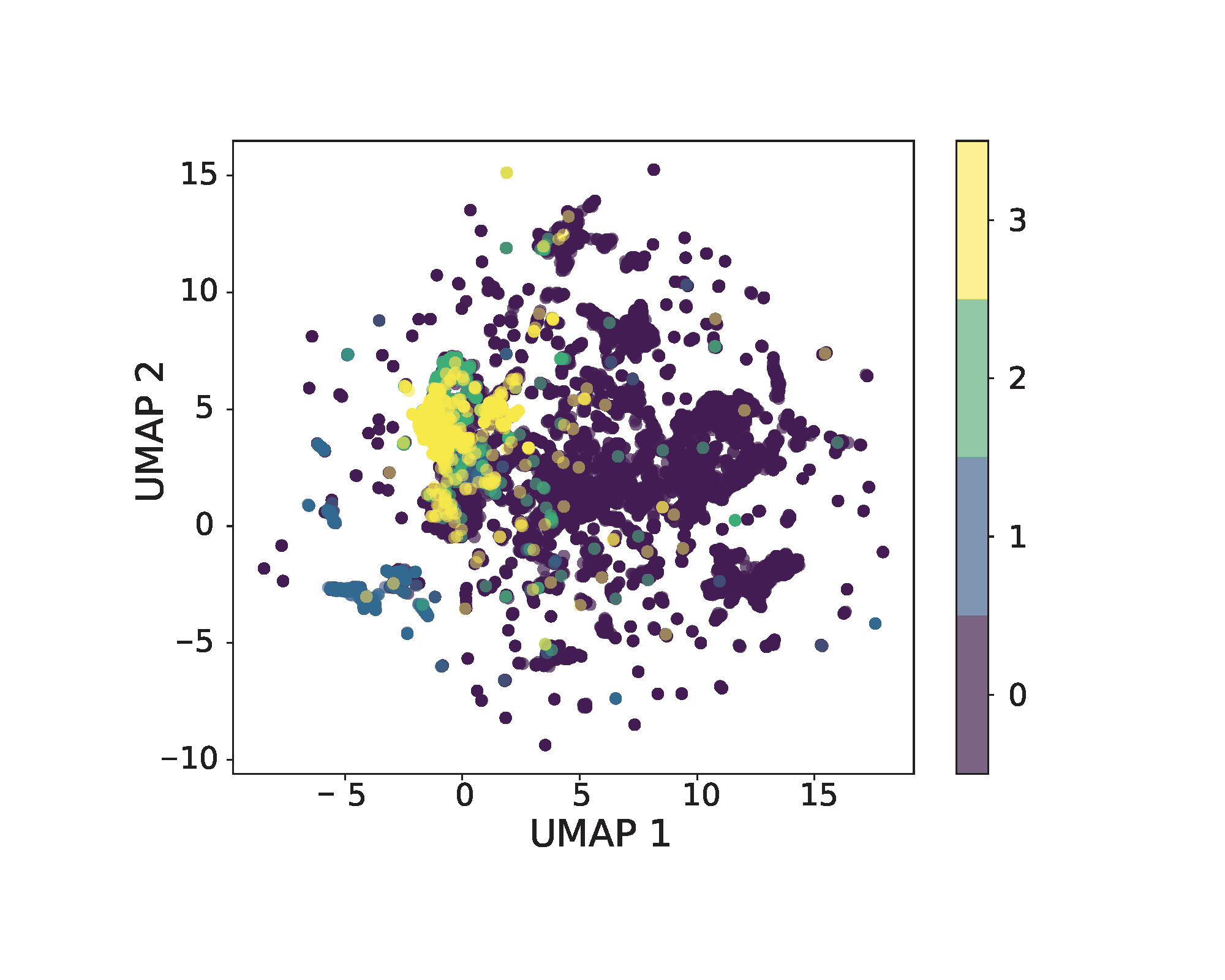}
        \caption{AG News (UMAP)}
        \label{fig:ag_news_umap}
    \end{subfigure}
    \hfill
    \begin{subfigure}[t]{0.48\columnwidth}
        \includegraphics[width=\textwidth, trim=80 60 115 50, clip]{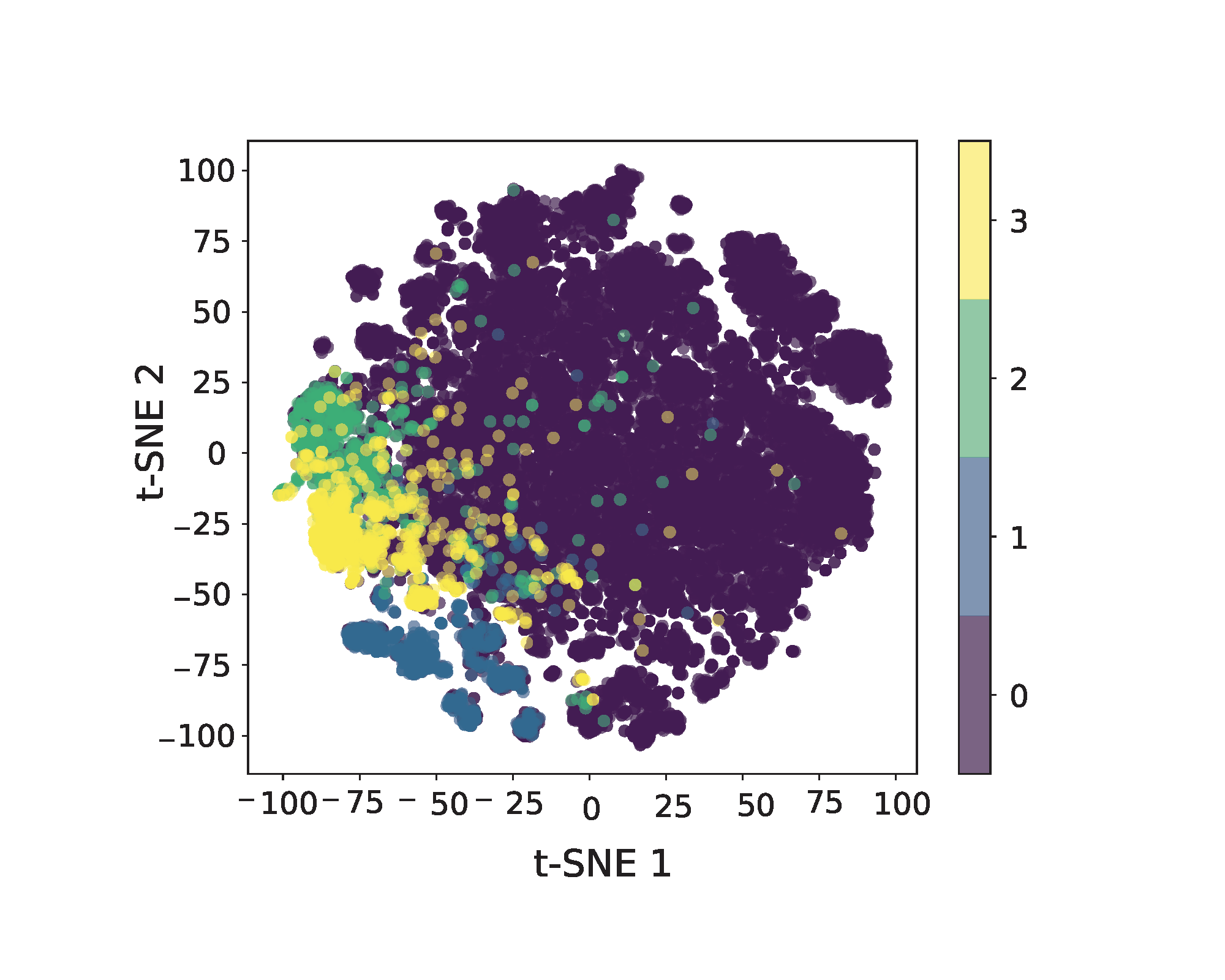}
        \caption{AG News (t-SNE)}
        \label{fig:ag_news_tsne}
    \end{subfigure}
    
    \begin{subfigure}[t]{0.48\columnwidth}
        \includegraphics[width=\textwidth, trim=80 60 115 50, clip]{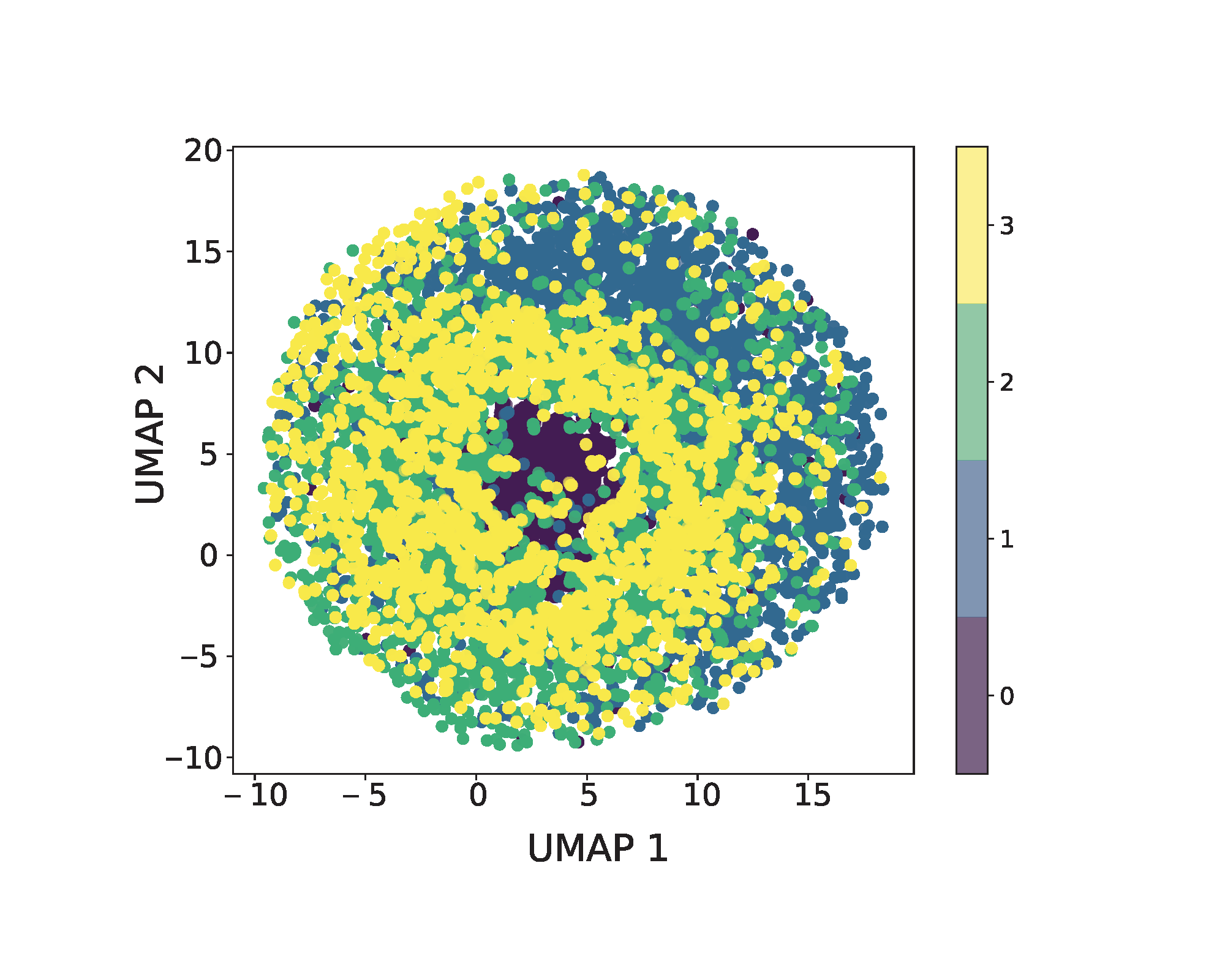}
        \caption{AG News (UMAP, SMOTE)}
        \label{fig:ag_news_umap_smote}
    \end{subfigure}
    \hfill
    \begin{subfigure}[t]{0.48\columnwidth}
        \includegraphics[width=\textwidth, trim=80 60 115 50, clip]{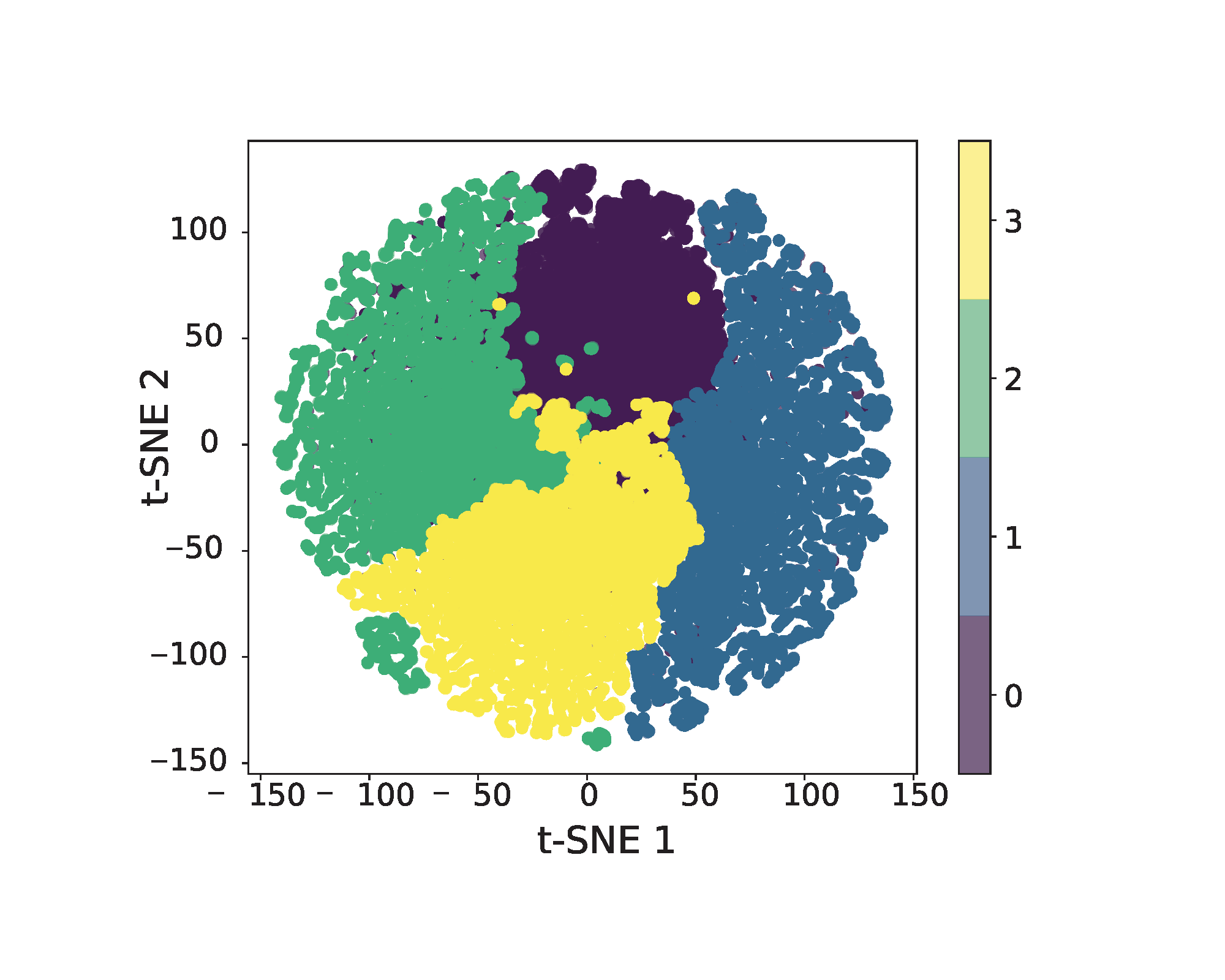}
        \caption{AG News (t-SNE, SMOTE)}
        \label{fig:ag_news_tsne_smote}
    \end{subfigure}

    \caption{
        Visualization of synthesized samples in the AG News dataset.
    }
    \label{fig:tsne_umap_agnews_plots}
        \vspace{-4mm}
\end{figure}

\begin{figure}[t] 
\captionsetup{font=footnotesize}
\centering
    \begin{subfigure}[t]{0.48\columnwidth} 
        \includegraphics[width=\textwidth, trim=70 60 115 50, clip]{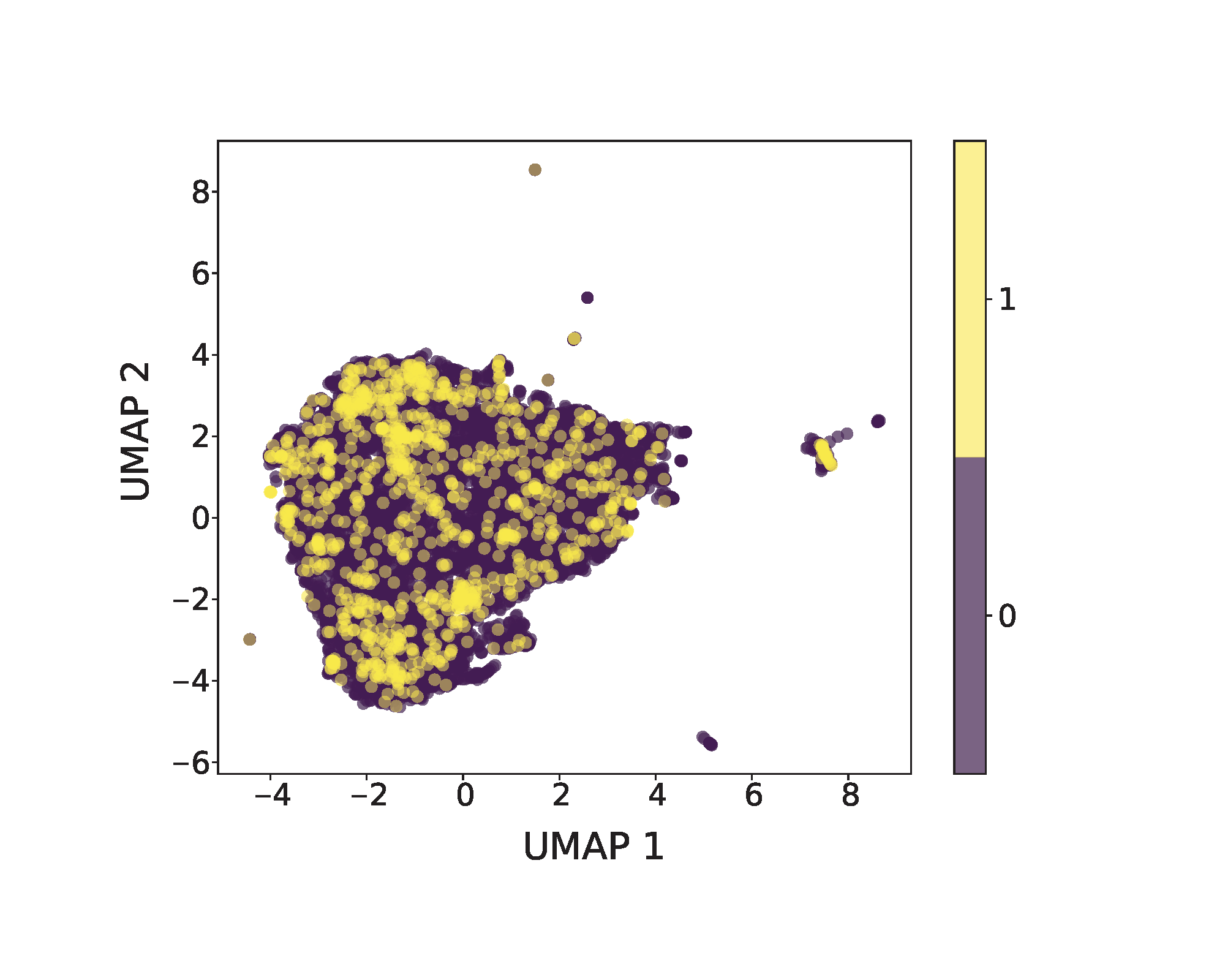}
        \caption{IMDB (UMAP)}
        \label{fig:imdb_umap}
    \end{subfigure}
    \hfill
    \begin{subfigure}[t]{0.48\columnwidth}
        \includegraphics[width=\textwidth, trim=70 60 115 50, clip]{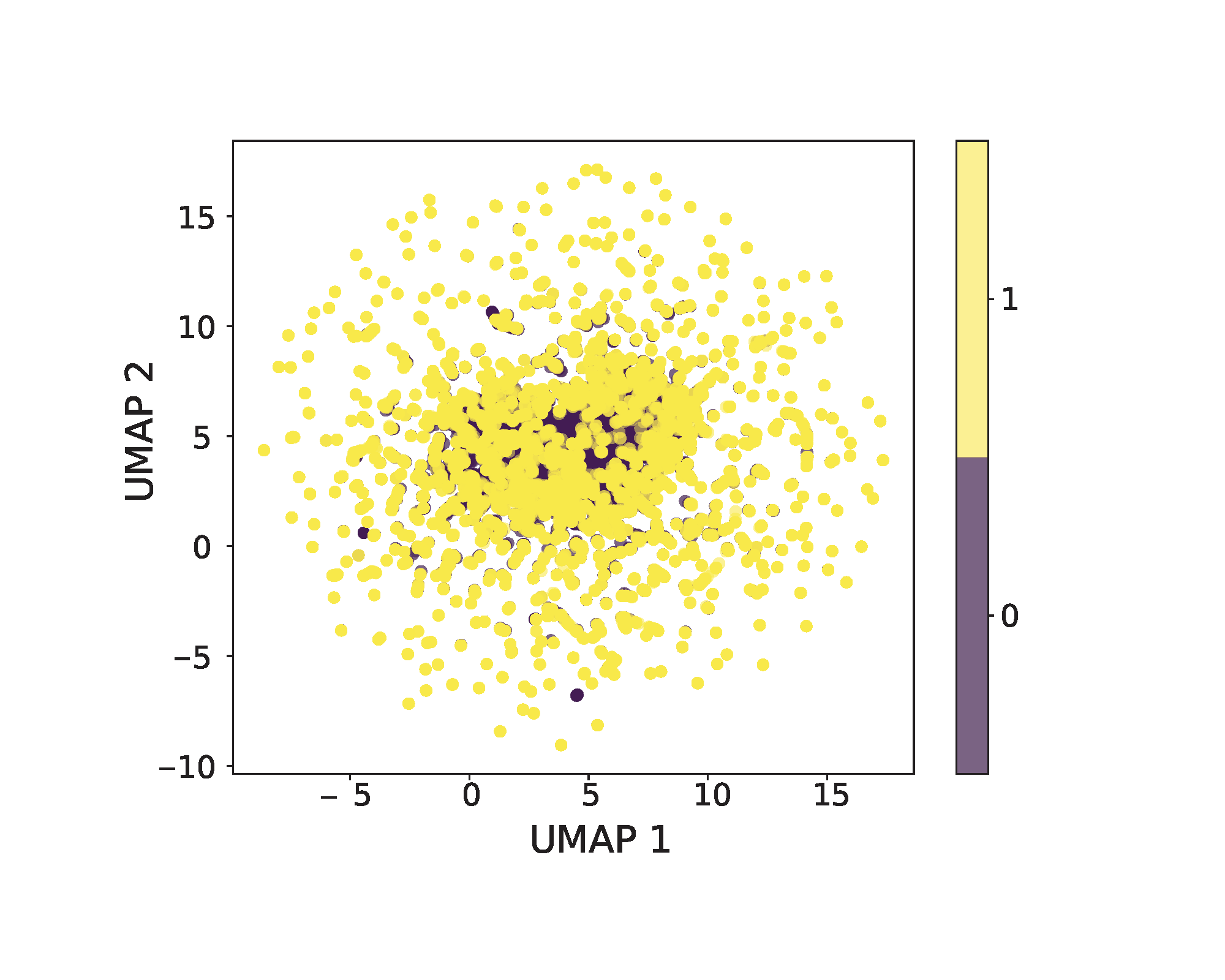}
        \caption{IMDB (UMAP, SMOTE Upsampled)}
        \label{fig:imdb_umap_smote}
    \end{subfigure}
    \vspace{0.3cm}
    \begin{subfigure}[t]{0.48\columnwidth}
        \includegraphics[width=\textwidth, trim=70 60 115 50, clip]{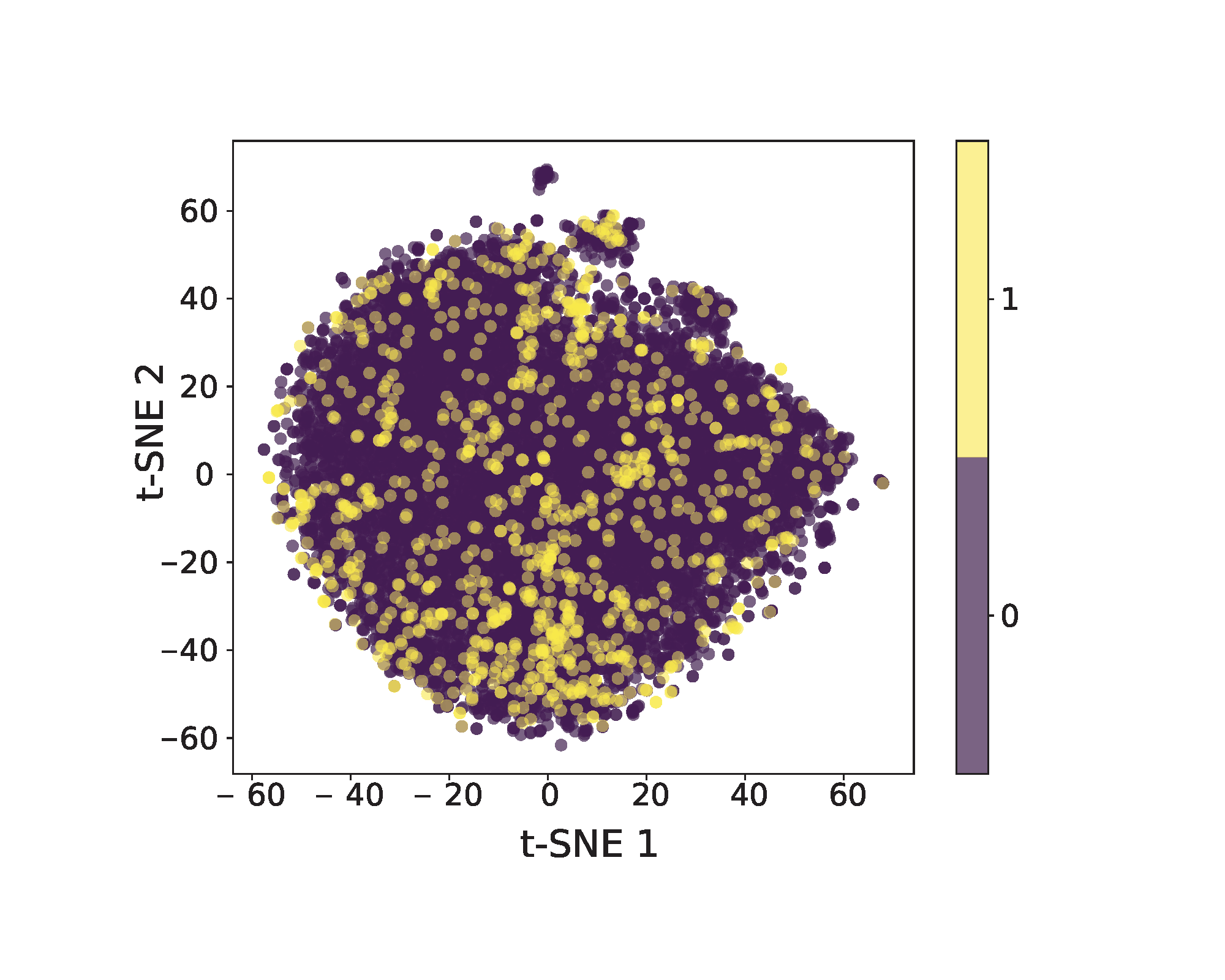}
        \caption{IMDB (t-SNE)}
        \label{fig:imdb_tsne}
    \end{subfigure}
    \hfill
    \begin{subfigure}[t]{0.48\columnwidth}
        \includegraphics[width=\textwidth, trim=70 60 115 50, clip]{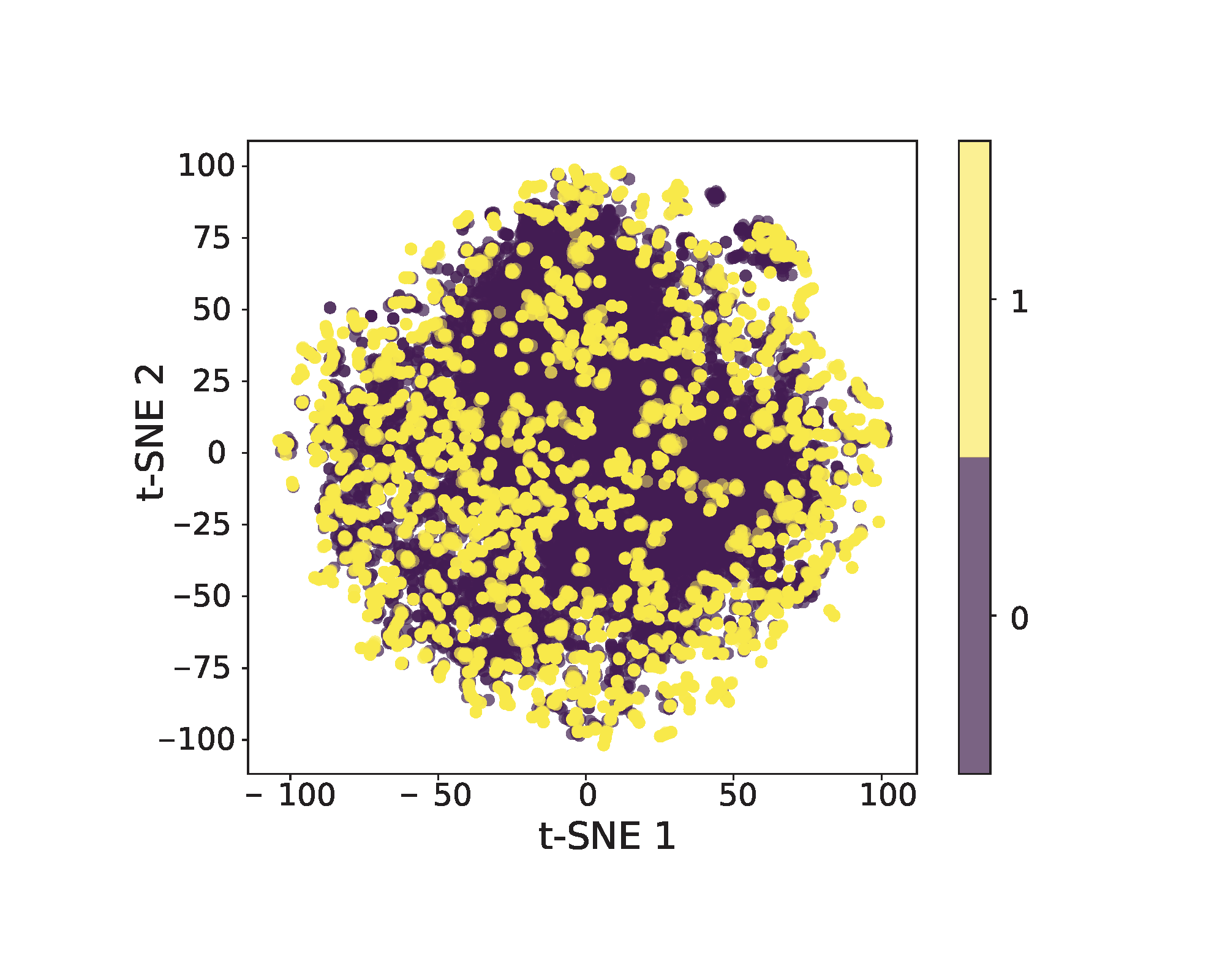}
        \caption{IMDB (t-SNE, SMOTE Upsampled)}
        \label{fig:imdb_tsne_smote}
    \end{subfigure}

    \caption{
        Visualization of synthesized samples in the IMDB dataset.
    }
    \label{fig:tsne_umap_imdb_plots}
        \vspace{-4mm}
\end{figure}

\begin{figure}[t]
\captionsetup{font=footnotesize}
\centering
    \begin{subfigure}[t]{0.24\textwidth} 
        \includegraphics[width=\textwidth, trim=70 60 115 50, clip]{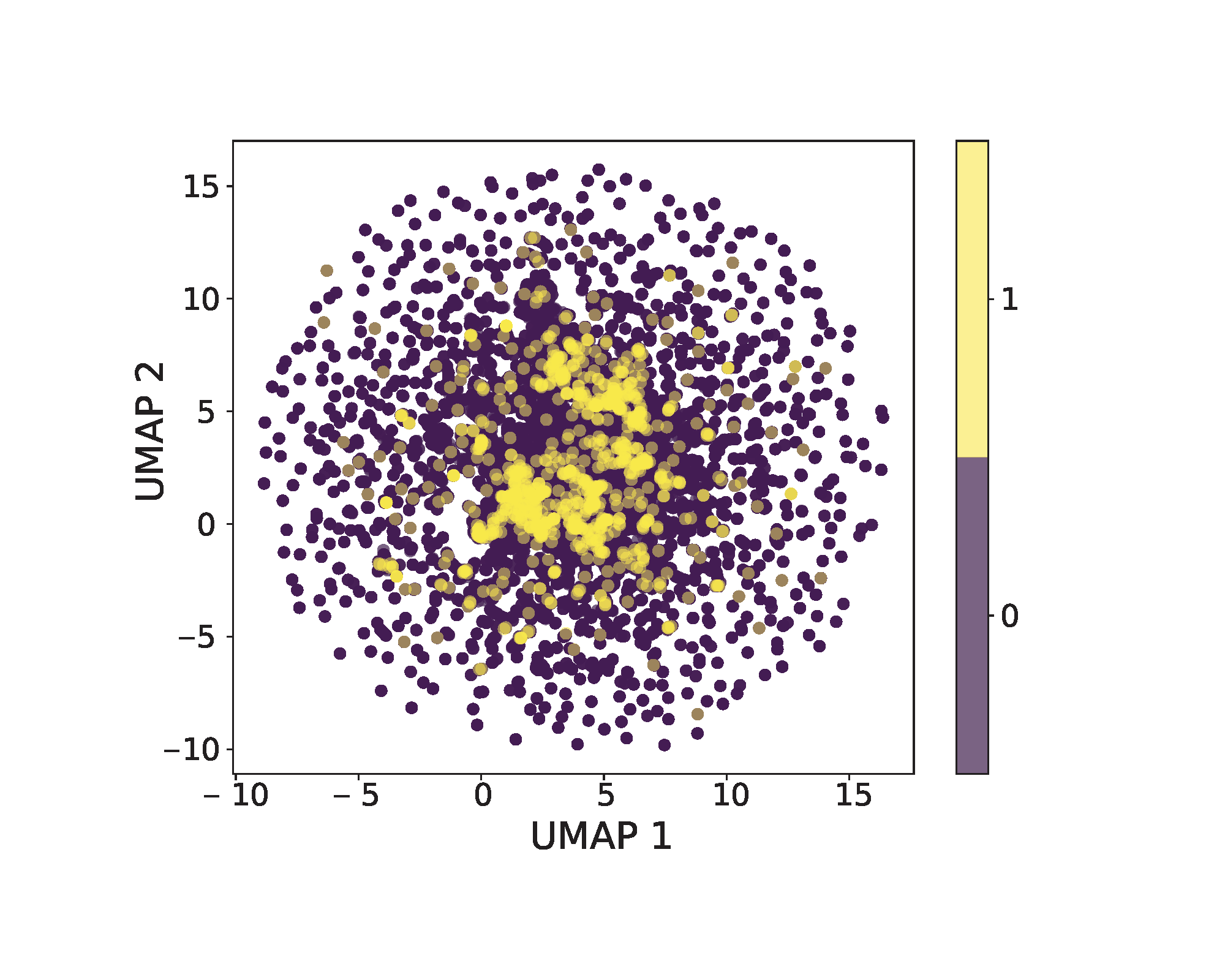}
        \caption{SST-2 (UMAP)}
        \label{fig:sst2_umap}
    \end{subfigure}
    \hfill
    \begin{subfigure}[t]{0.24\textwidth}
        \includegraphics[width=\textwidth, trim=70 60 115 50, clip]{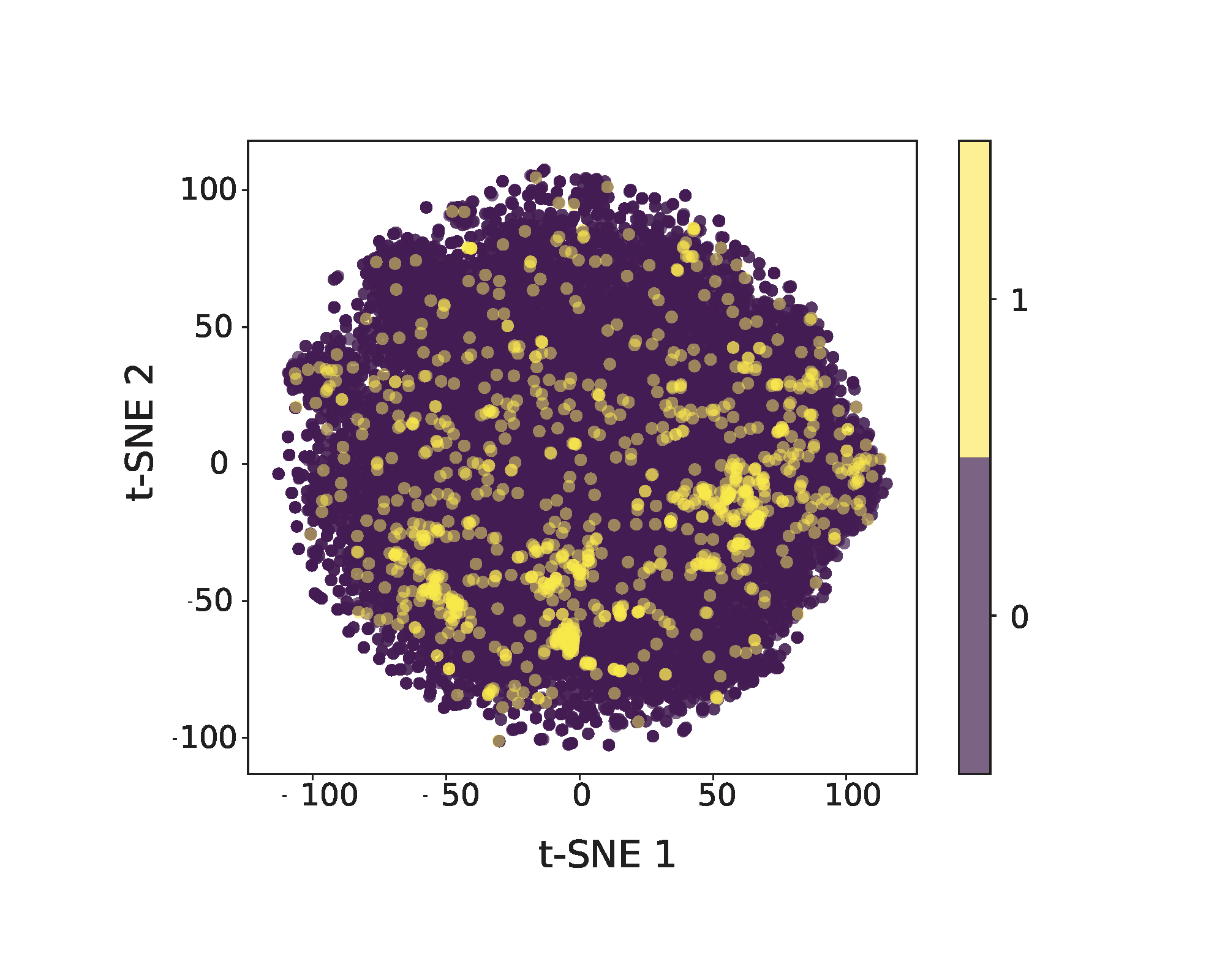}
        \caption{SST-2 (t-SNE)}
        \label{fig:sst2_tnse}
    \end{subfigure}
    \hfill 
    \begin{subfigure}[t]{0.24\textwidth}
        \includegraphics[width=\textwidth, trim=70 60 115 50, clip]{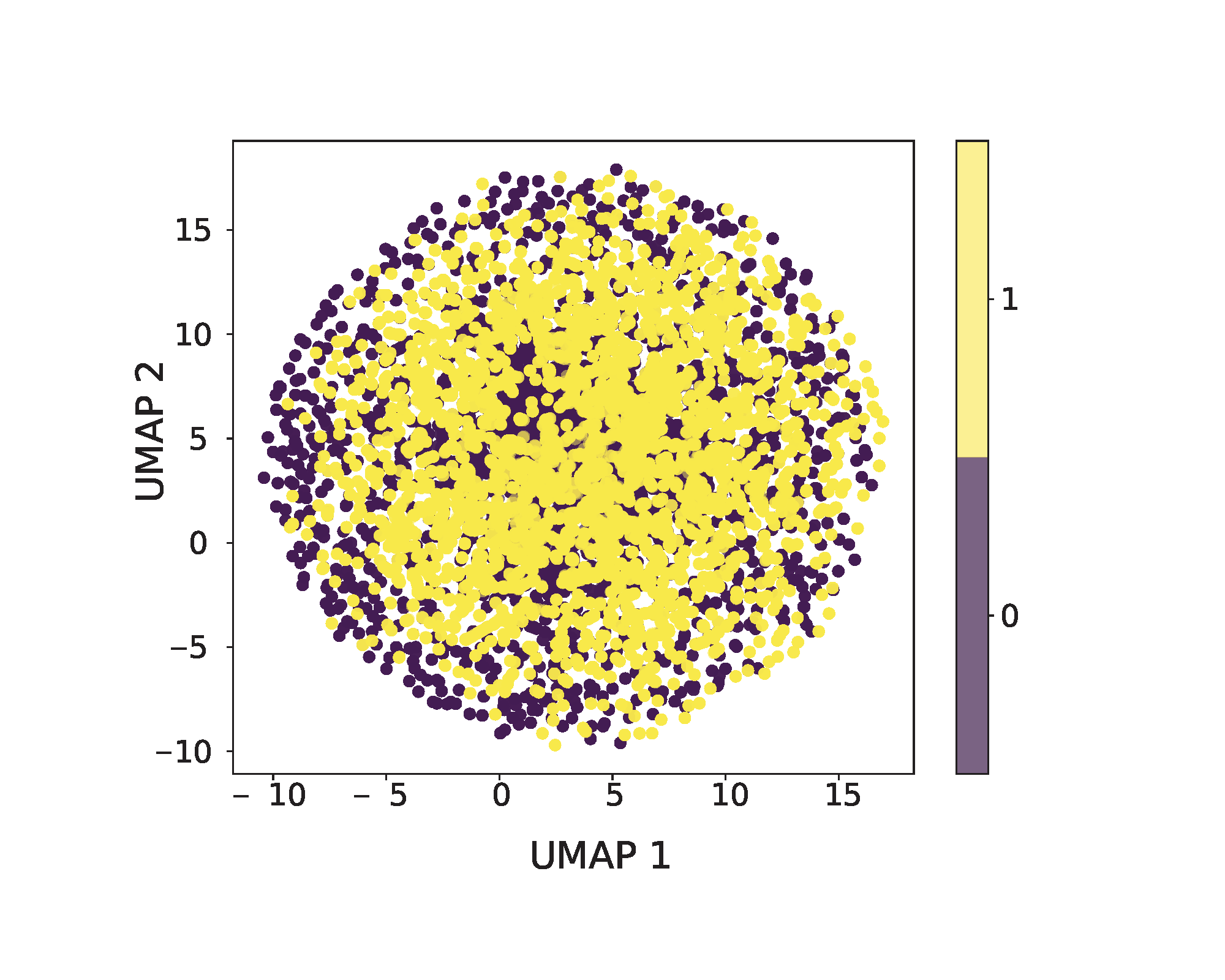}
        \caption{SST-2 (UMAP, SMOTE Upsampled)}
        \label{fig:sst2_umap_smote}
    \end{subfigure}
    \hfill 
    \begin{subfigure}[t]{0.24\textwidth}
        \includegraphics[width=\textwidth, trim=70 60 115 50, clip]{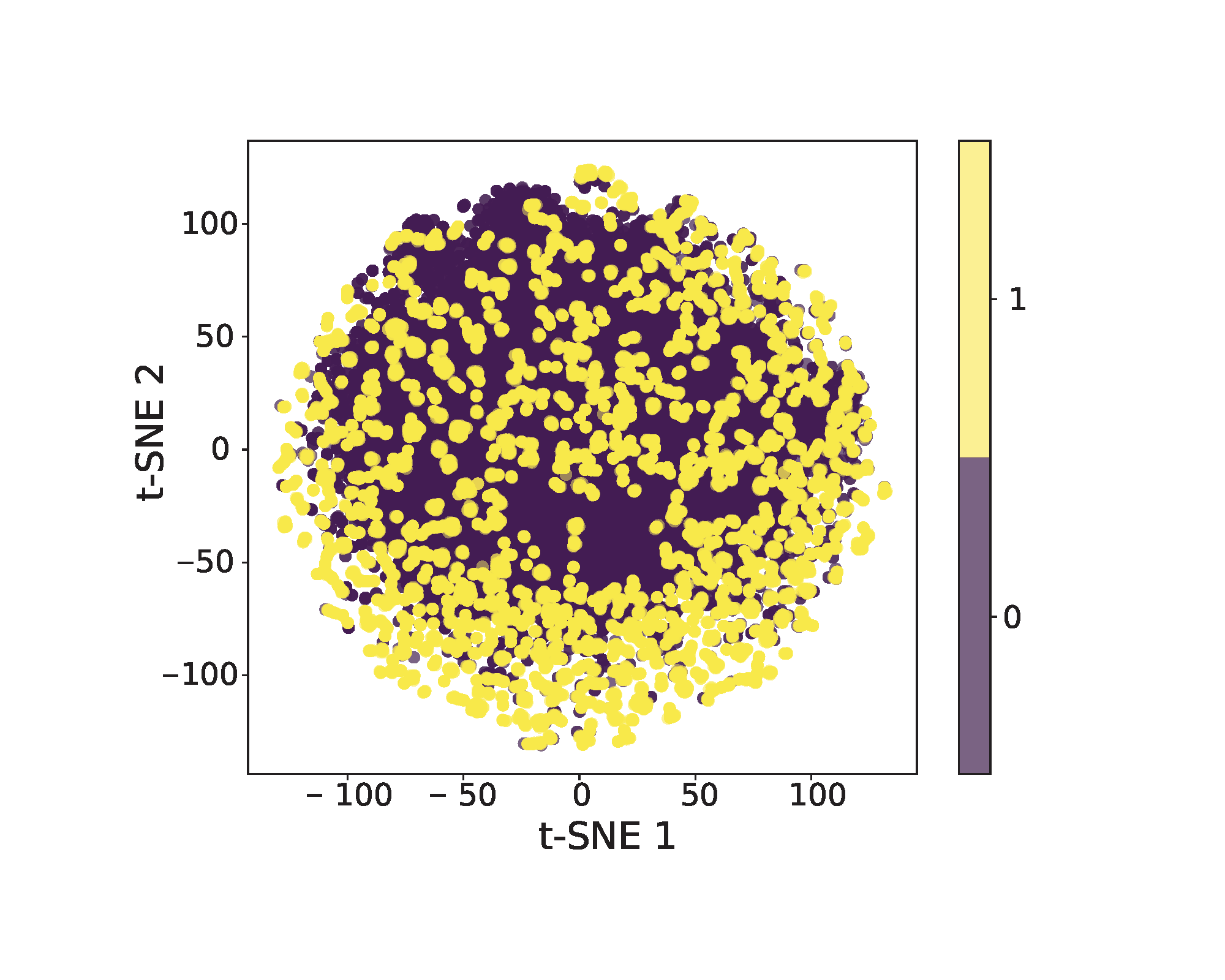}
        \caption{SST-2 (t-SNE, SMOTE Upsampled)}
        \label{fig:sst2_tnse_smote}
    \end{subfigure}
    \caption{
        Visualization of synthesized samples in the SST-2 dataset.
    }
    \label{fig:tsne_umap_sst2_plots}
    \vspace{-4mm}
\end{figure}


The IMDB and SST-2 datasets, being binary classification tasks, mitigate class imbalance to some extent. However, on the multiclass AG News dataset with four classes, SMOTE outperformed VAE in accuracy, despite less pronounced overall improvement. This demonstrates SMOTE's versatility and effectiveness across classification tasks.  
The results highlight SMOTE's ability to enhance model performance across sample sizes, emphasizing the importance of appropriate augmentation techniques for imbalanced datasets.

\subsection{Embedding Space Analysis}


Figure \ref{fig:tsne_umap_agnews_plots} presents the 2D projection of embeddings using uniform manifold approximation and projection (UMAP)~\cite{McInnes2018} and t-distributed stochastic neighbor embedding (t-SNE )~\cite{vandermaaten08a} from the AG News dataset, comparing results with and without SMOTE oversampling. Similar visualizations are provided for the IMDB and SST-2 datasets in Figures~\ref{fig:tsne_umap_imdb_plots} and \ref{fig:tsne_umap_sst2_plots}, which shows performance of the proposed method in synthesizing samples in the embedding space.

\section{Conclusion}
This study evaluated SMOTE and VAE for handling class imbalance in binary (IMDB, SST-2) and multiclass (AG News) text classification tasks. By generating synthetic data for minority classes and comparing performance, we found that SMOTE consistently outperformed VAE in binary tasks and showed better results in multiclass scenarios, though improvements were less pronounced.  
These findings highlight SMOTE as a reliable method for addressing class imbalance and underscore the importance of effective data augmentation in improving model performance on imbalanced datasets.


{\small
\bibliographystyle{unsrt}
\bibliography{main}
}

\end{document}